\documentclass[twoside,11pt]{article}
\usepackage[preprint]{jmlr2e}

\usepackage{amssymb,amsmath,graphicx,url,times,fourier}
\DeclareMathOperator{\arctantwo}{arctan2}

\usepackage{hyperref}
\usepackage{appendix}
\usepackage{multirow}
\usepackage{xcolor}
\usepackage{lineno}

\usepackage[nonumberlist]{glossaries}
\makeglossaries

\usepackage{lastpage}
\ShortHeadings{}{Torus embeddings}
\firstpageno{1}

\title{Torus Embeddings}
\author{Dan Stowell \\ \addr Tilburg University\\Tilburg\\The Netherlands \AND \addr Naturalis Biodiversity Centre\\Leiden\\The Netherlands}

\editor{Editor TBC}

\begin{document}
\maketitle

\begin{abstract}%   <- trailing '%' for backward compatibility of .sty file
Many data representations are vectors of continuous values.
In particular, deep learning embeddings are data-driven representations, typically either unconstrained in Euclidean space, or constrained to a hypersphere.
These may also be translated into integer representations (quantised) for efficient large-scale use.
However, the fundamental (and most efficient) numeric representation in the overwhelming majority of existing computers is integers \textit{with overflow}---and vectors of these integers do not correspond to either of these spaces, but instead to the topology of a (hyper)torus.
This mismatch can lead to wasted representation capacity.
Here we show that common deep learning frameworks can be adapted, quite simply, to create representations with inherent toroidal topology.
We investigate two alternative strategies, demonstrating that 
a normalisation-based strategy leads to training with desirable stability and performance properties, comparable to a standard hyperspherical L2 normalisation.
We also demonstrate that a torus embedding maintains desirable quantisation properties.
The torus embedding does not outperform hypersphere embeddings in general, but is comparable, and opens the possibility to train deep embeddings which have an extremely simple pathway to efficient `TinyML' embedded implementation.
\end{abstract}


%%%%%%%%%%%%%%%%%%%%%%%%%%%%%%%%%%%%%%%%%%%%%%%%%%%%%%%%%%%%%%%%%%%%%%%%%%%%%%%%%%%%%%%%%
%%%%%%%%%%%%%%%%%%%%%%%%%%%%%%%%%%%%%%%%%%%%%%%%%%%%%%%%%%%%%%%%%%%%%%%%%%%%%%%%%%%%%%%%%
%%%%%%%%%%%%%%%%%%%%%%%%%%%%%%%%%%%%%%%%%%%%%%%%%%%%%%%%%%%%%%%%%%%%%%%%%%%%%%%%%%%%%%%%%
\section{Introduction}

Data representations are designed once, and reused many times.
This remains true of the data-driven representations now widespread in the era of deep learning (DL) foundation models: training such a model establishes the parameters for a function that may be invoked many millions of times to project new data points.
A common use-case for such high-dimensional representations is distance-based search, and for large-scale use this search must be as efficient as possible.
In deep learning, the combination of ``standard'' large-scale DL training (e.g. for supervised classification), followed by distance-based inference in the representation space that has been created, is a powerful and increasingly popular paradigm \citep{Oquab:2024}%DINOv2
\citep{Luo:2023}%that fewshot NNC paper
.
For example it has recently been shown that for few-shot learning, distance-based search (kNN) can a strongly competitive and efficient query method, compared against algorithms more specialised for the task, even when applied to embedding spaces that have been trained for other types of query \citep{Luo:2023}.

This increases the importance of high-dimensional distance-based search in vector spaces.
Even if the training of a representation is computationally heavy, the overarching need is for the produced representation to be very efficient in use, and suitable for purposes such as similarity search or range-based retrieval \citep{Jegou:2011}%PQ paper
.
The efficiency is affected by algorithmic choices, but also by their implementation in hardware.
Even the implementation of number formats matters; hence the investigation of floating-point and fixed-point representations for deep learning implementations
\citep{Burgess:2019,Dettmers:2023,Rouhani:2023}. % BF16
While novel number formats can be supported by novel processor tooling, an alternative route is for data representations that make the best use of the most lo-fi and fundamental representations that general-purpose computers can offer.
It is highly desirable to obtain deep representations (embeddings) that can be implemented efficiently across all common computing platforms.
While IEEE floating-point is widespread, and is the starting point for most deep learning, integer processing can be even more efficient and even more widespread---which motivates much work on discrete encodings, and quantised neural networks \citep{Jacob:2018,%QAT
Menghani:2023%survey paper
}.
Further: while GPUs, TPUs and NPUs can perform high-dimensional calculations efficiently, their global availability is dwarfed by traditional CPUs by many orders of magnitude. There is a tradeoff between two different strategies for large-scale deployment of embeddings: the efficiency of using novel hardware, versus the resource-saving circularity and the inherent scalability of reusing existing widespread platforms \citep{Knowles:2022}.

In the field of DL, recent work highlights that hyperspherical spaces have useful properties for data representation, compared against unbounded Euclidean space \citep{
Wang:2020,%contr using uniformity on the hypersphere
Trosten:2023,%hubs and hyperspheres
LinSong:2020,%minimising hyperspherical energy
Wang:2017%normface
}, and in particular that they can be trained well by deep neural networks \citep{Wang:2017,Wang:2020}.
However, points on a hypersphere are not a natural fit to low-bitrate integer representations, since there is no direct coordinate mapping for the hypersphere.
Note that vectors constrained to a hypersphere give coordinate values whose distributions are distinctly different from both uniform and normal distributions---raising the question of how best to quantise such a representation.
Instead, coding-decoding schemes are used \citep{Sablayrolles:2018}%koleo
.
There may be good reasons to choose something unorthodox.

In the present work, instead of changing the numeric representation
we ``flip the problem'' and choose a topological space that naturally transfers to the most basic number representation on computers.
We ask how to bring the high- and low-resolution representation spaces into closer aligment, via a different choice of topological space.
We argue for the (hyper)torus as the natural choice for representations in this domain.
We explore toroidal representations through a simple deep learning experiment.

Following the principles of permacomputing, we target representations that can be used on a wide range of hardware including both modern and old CPU architectures
\citep{Mansoux:2023,%permacomputing
Jacob:2018%QAT
}. For this, the most common and most efficient data representation is 8-bit integers with overflow arithmetic. We demonstrate that these correspond to a torus topology and show how a modern training pipeline can produce toroidal representations directly. Torus embeddings are easy to integrate into many deep learning pipelines, and perform well at low bitrate.

\begin{table}[t]
\centering
\begin{tabular}{ccccc}
Original space   &   Transform   &   Space   &   Extrinsic dimension   & Intrinsic dimension \\
\hline
\multirow{5}{*}{\raisebox{-0.3\height}{\includegraphics[width=1.3cm]{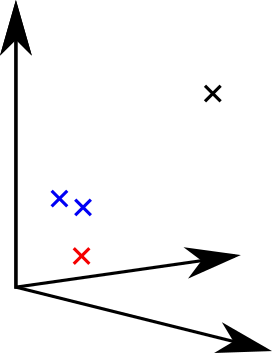}}}           &  $\xrightarrow{L_2}$         &   \raisebox{-0.5\height}{\includegraphics[width=1.1cm]{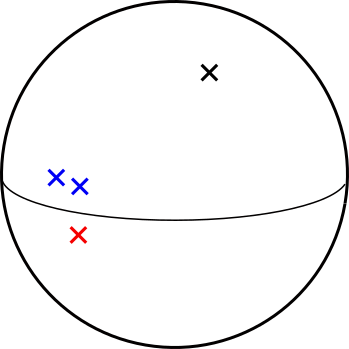}} Hypersphere  &   $D$             & $D-1$ \\
\cline{3-5}
           &  $\xrightarrow{\mod{}}$      &   \raisebox{-0.3\height}{\includegraphics[width=1.2cm]{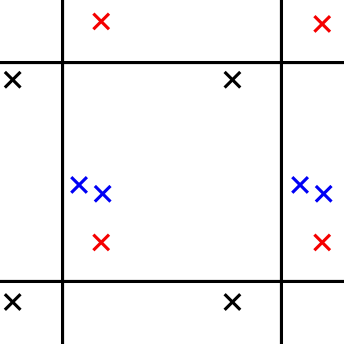}} Flat torus \hspace{5mm}  &   $D$             & \multirow{2}{*}{$D$} \\
           &  $\xrightarrow{[\sin,\cos]}$ &   \raisebox{-0.3\height}{\includegraphics[width=1.3cm]{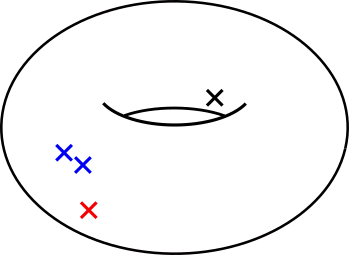}} Clifford torus   &   $2D$            & \\
\cline{3-5}
           &  $\xrightarrow{\arctantwo}$  &   \raisebox{-0.3\height}{\includegraphics[width=1.2cm]{pointsinspaces_flattorus}} Flat torus \hspace{5mm} &   $D/2$           & \multirow{2}{*}{$D/2$} \\
           &  $\xrightarrow{L_{2p}}$      &   \raisebox{-0.3\height}{\includegraphics[width=1.3cm]{pointsinspaces_roundtorus}} Clifford torus   &   $D$             & \\
\hline
\end{tabular}
  \caption{Normalisation projections for deep learning embeddings. Starting from a `raw' unconstrained representation of dimension $D$, $L_2$ normalisation is just one possible projection. We consider two approaches to normalisation that create hypertoroidal representations: each has two projections, one leading to the `flat square' torus and one leading to the unit-norm `Clifford' torus. See text for further explanation.}
  \label{fig:topo_overview}
\end{table}

Contributions:

\begin{itemize}
\item   We describe two methods for training embeddings with hypertoroidal geometry, both of which are implemented as simple modifications in a standard deep learning pipeline. We demonstrate that one method is more stable to train.
\item   We show that KoLeo regularisation \citep{Sablayrolles:2018} is beneficial to train hypertoroidal representations in some circumstances.
\item   We investigate the quantisation properties of toroidal and hyperspherical deep representations. We argue that toroidal representations are a good match for quantised representations.
Our results show that while our full-resolution results do not outperform a hyperspherical embedding, our torus representation retains the strongest fidelity in some settings under extremely low-bitrate quantisation.
However we also observe that product quantisation (PQ) yields good compression results for both the hypertorus and the standard hypersphere, for all except the absolute lowest bit rates.
This exposition also yields some observations on why hyperspherical embeddings are generally stable to train.
\end{itemize}

In this paper, we work in the paradigm of training a deep neural network using floating-point values, and then translating the system to a low-power integer platform. We do not consider the quantisation of the network parameters, which has been extensively covered elsewhere\citep{Jacob:2018,%QAT
Menghani:2023%survey paper
}. Instead we focus on the topology of the final representation. We consider how best to create a multidimensional embedding space, to be used for distance-based retrieval.

Before describing the specifics of our method, we first consider the relevant topological issues and terminology.

%%%%%%%%%%%%%%%%%%%%%%%%%%%%%%%%%%%%%%%%%%%%%%%%%%%%%%%%%%%%%%%%%%%%%%%%%%%%%%%%%%%%%%%%%
\section{Topologies of Deep Embedding Spaces}

In machine learning and especially representation learning, good embedding spaces are compact, bounded, metric, and homogeneous (all points are equivalent), the latter implying they must be borderless.
In many works, embeddings are in $R^D$. However, this is neither compact nor bounded. This can lead to problems in training, such as divergence, commonly addressed by regularisation \citep{Goodfellow:2016textbook}. For around 10 years (since papers such as \citet{Wang:2017}%NORMFACE
), a popular alternative is L2-normalised embeddings, whose points thus lie on a hypersphere. The hypersphere has desirable properties and good performance in practice
\citep{
Wang:2020,%contr using uniformity on the hypersphere
Trosten:2023,%hubs and hyperspheres
LinSong:2020,%minimising hyperspherical energy
Wang:2017%normface
}.

Hypersphere normalisation should not be considered as merely an engineering tweak for stability/efficiency: it is a change of topology for the representation space, and thus determines characteristics of representations.
The hypersphere is a common embedding space, but others advocate for $R^D$, and the latter is common in Transformers (but see \citet{Loshchilov:2024}%nGPT
). The choice of embedding space is often unexamined. There are good arguments in the literature, but not all univocal.

The hypersphere, and also unconstrained $R^D$, are inefficient in terms of their numeric representation in computers, in the sense that a representation would ideally tend to distribute data uniformly across all representable vectors---indeed many optimisation schemes set this as an explicit goal \citep{Sablayrolles:2018,Wang:2020}.
The hypersphere especially is not well-matched to the underlying representation---neither integers nor floating-points---since all data is confined to the spherical submanifold, not the entire space.
	Workarounds such as spherical quantisation schemes acheve data compression at the cost of extra computation.

On the vast majority of CPU designs, unsigned integer bytes (``uint8'') are the simplest representation%, and overflow (or non-saturating) arithmetic is the cheapest and fastest operation
. This is true for older as well as current instruction set architectures (ISAs) such as RISC-V \citep{Intel:1979,%x86
RISCV:2019}%RISC-V
. Other bit widths are often supported too, such as uint16 and uint32.
Note also that on many platforms, the fundamental integer addition operation is non-sturating with a wrap-around behaviour that corresponds directly to the modulo operation \citep[section 2.4]{RISCV:2019}%RISC-V Manual Volume 1: the unprivileged ISA -- section 2.4
. Modulo addition of two integers is then performed with two atomic instructions: the add itself, plus ``clear carry'' or equivalent to discard the carry bit. Non-modulo addition is in general slower, since overflow must be checked and then handled. By contrast, modulo addition invokes no program branches and so will not stall a CPU pipeline.
This is reflected for example in C/C++, where unsigned integers are defined with modulo wrapping as the intended behavior.
A scalar integer space with modulo behaviour corresponds to a ring or circular topology.

Combining these considerations: a vector of atomic integers, plus arithmetic operations having overflow (modulo) behaviour, naturally define a space with the topology of a torus or hypertorus. (Henceforth, we will use the term `torus' generally to include any higher-dimensional hypertorus; cf.\ Table \ref{tbl:glossary}.) Specifically it correponds to a \textit{flat, square} torus, in which each dimension has the same size and zero curvature \citep{Courbe:2024}. The set of all possible vectors is a set distributed as a regular grid covering all parts of the torus with equal density. 
Representations which use this topology should therefore provide opportunities to translate well to low-level representations, e.g.\ by quantisation.
This suggests that a very wide range of hardware devices can support efficient operations in toroidal spaces---and thus, that it is desirable to be able to produce toroidal data embeddings.

Happily, hypertoroidal spaces also have many desirable properties for deep learning representations: they are compact, bounded, metric and homogenous.
High-dimensional torus spaces exhibit the same general curse-of-dimensionality characteristics as do high-dimensional spherical spaces (Figure \ref{fig:plot_torusdist_distribs}).

In the remainder of this paper, we consider how to perform training and inference of toroidal representations within deep learning frameworks. Sections 3 and 4 respectively cover training and inference. Experiments (Section 5) illustrate the method in practice, and compare the characteristics of torus embeddings against standard hyperspherical representations.

%%%%%%%%%%%%%%%%%%%%%%%%%%%%%%%%%%%%%%%%%%%%%%%%%%%%%%%%%%%%%%%%%%%%%%%%%%%%%%%%%%%%%%%%%
%%%%%%%%%%%%%%%%%%%%%%%%%%%%%%%%%%%%%%%%%%%%%%%%%%%%%%%%%%%%%%%%%%%%%%%%%%%%%%%%%%%%%%%%%
%%%%%%%%%%%%%%%%%%%%%%%%%%%%%%%%%%%%%%%%%%%%%%%%%%%%%%%%%%%%%%%%%%%%%%%%%%%%%%%%%%%%%%%%%
\section{Training Torus Representations}

\begin{table}[t]

\newglossaryentry{flatsquaretorus}{
  name={Flat square (hyper)torus},
  description={An embedding which has a (hyper)torus geometry, represented on D dimensions with axis having a flat S1 geometry. In this work these are represented as values in $[0, 1]^D$}
}

\newglossaryentry{cliffordtorus}{
    name={Clifford (hyper)torus},
    description={Any embedding which has a (hyper)torus geometry and in which all points have the same L2 norm in the ambient space}
}
\newglossaryentry{cliffordproj}{
    name={Clifford projection},
    description={A projection where each dimension goes from x to [sin x, cos x], thus one way to create a Clifford (hyper)torus}
}
\newglossaryentry{l2p}{
    name={Pairwise L2 normalisation ($L_{2p}$)},
    description={A projection in which each pair of dimensions is L2-normalised, without regard to the others; a second way to create a Clifford (hyper)torus}
}
\glsaddall
\printglossary

  \caption{Terminology used in this paper.}
  \label{tbl:glossary}
\end{table}

We first consider how toroidal spaces can be used within the straining of deep learning, using gradient descent and well-known loss functions.

We immediately note that using a toroidal space may constrain some choices of method.
For example, the cyclical topology means we cannot in general define a hyperplane in the flat torus space that partitions the data in two. For this and analogous reasons, it is not meaningful to apply sigmoid or softmax nonlinearities in the space.
However, distance-based algorithms can be applied straightforwardly, such as triplet or contrastive learning. Contrastive learning has recently been shown to be a powerful method, both for self-supervised and supervised learning \citep{Chen:2020simclr,%simCLR
Khosla:2020%supcon
}.
In our experiments, we will follow this line, using supervised contrastive learning (`SupCon') for distance-based training.

Yet there remains one further barrier.
In training a distance-based representation, a common requirement is to calculate all-pairs distances (or similarities) within a single batch.
Distance in a flat torus is a simple concept, but there are in fact many straight lines between any given pair of points, since each dimension can be traversed in one of two directions. To find the shortest (the geodesic) an algorithm must check each of these, and this must be done for each pair of points in a batch.
For all-pairs calculations in high-dimensional deep learning this would be prohibitively expensive.

The Clifford projection offers a solution to this dilemma.
The \textit{Clifford torus} is an embedding of the two-dimensional flat torus created by the mapping
\begin{equation}\label{eq:clifford2}
(x, y) \rightarrow \sqrt{\frac{1}{2}} (\sin x, \cos x, \sin y, \cos y)
\end{equation}
for the case that $x$ and $y$ represent coordinates in a flat square torus of size $2 \pi$. This is a \textit{distortion-free} mapping of the original torus space. % (Figure \ref{fig:torus_flat_clifford}).
In the Clifford embedding of the torus, all points lie at the same fixed distance from the origin. Then cosine distance is a natural and efficient distance measure, just as it is in hyperspherical representations,
and sidesteps the combinatorial issue.
(Trigonometric functions are only needed at training time, as will be explained.)
We will apply projections of this type during training, albeit our input spaces will usually be of dimension higher than 2.

We consider two different approaches to create Clifford hypertori, each having a corresponding projection to the flat square torus (Table \ref{fig:topo_overview}).
First, the Clifford projection \eqref{eq:clifford2} extended for input spaces of dimension $D$:

\begin{equation}\label{eq:cliffordD}
(x_1, ... x_D) \rightarrow \sqrt{\frac{1}{D}} (\sin x_1, \cos x_1, ... \sin y_D, \cos y_D)
\end{equation}

Second, we note an alternative strategy.
The Clifford hypertorus is a subspace of a hypersphere, as is clear from the fact that all points have a fixed L2 norm. Points on the Clifford hypertorus are further constrained such that each \textit{pair} of dimensions (each two-dimensional subspace) also has a fixed L2 norm.
Thus we can define a projection similar to the well-known L2 normalisation, but applied pairwise to dimensions, which will map data into the Clifford hypertorus:

\begin{equation}\label{eq:L2p}
(x_1, x_2, ... x_{D-1}, x_D) \rightarrow \sqrt\frac{2}{D} \left(%
\frac{x_1}{||(x_1, x_2)||_2},%
\frac{x_2}{||(x_1, x_2)||_2},%
... %
\frac{x_{D-1}}{||(x_{D-1}, x_D)||_2},%
\frac{x_D}{||(x_{D-1}, x_D)||_2}%
\right)
\end{equation}
We will refer to this as pairwise L2 projection, or L2p for short.

Both of the above projections map arbitrary data onto the Clifford torus.
Note that we use the terms ``Clifford torus'' or ``Clifford space'' for any space having this geometry (i.e.\ with each pair of dimensions having constant L2 norm), even if not produced by the Clifford projection \eqref{eq:cliffordD} (cf.\ Table \ref{tbl:glossary}).
For any such space, \eqref{eq:L2p} is an identity mapping.
However, the projections \eqref{eq:cliffordD} and \eqref{eq:L2p} are quite different mappings (Figure \ref{fig:rainbowquant}) and this may have important consequences in practice, not least for gradient-based backpropagation of errors when training a representation.

There are also differences in the handling of dimensionality.
The extrinsic and intrinsic dimension of a representation can change as a result of the projections here described (Table \ref{fig:topo_overview}).
Standard hypersphere normalisation does not alter the extrinsic dimension (of the ambient space), but reduces the intrinsic dimension (the degrees of freedom) to $D-1$.
The L2p projection \eqref{eq:L2p} is closely related: it does not affect the extrinsic dimension, but halves the intrinsic dimension to $D/2$.
By contrast, the Clifford projection \eqref{eq:cliffordD} doubles the extrinsic dimensionality of our space, but the intrinsic dimension remains the same.
In the present work we use $D$ to represent the dimension of the input data before projection. These differences in the resulting dimensionalities should be borne in mind.

In the Clifford space, cosine distance is a natural distance function and efficient to implement.
It is not directly equivalent to any $L_p$-norm in the corresponding flat torus, but has very similar behaviour to $L_1$ and $L_2^2$ norms (Appendix \ref{cosinelikeLp}).

Projection onto a Clifford torus can be implemented as a nonlinearity in almost any modern deep learning network architecture, simply by using the projections described above in place of standard L2 normalisation $x \gets x/||X||_2$.
In particular, the L2p pariwse normalisation of \eqref{eq:L2p} can be applied very efficiently in frameworks such as PyTorch and TensorFlow, by creating a `view' of the tensor with an altered `shape' but without having to duplicate the tensor data in memory.

The Clifford hypertorus is a subspace of a hypersphere, and thus shares many properties with the hypersphere.
This includes the relevance of cosine distance as a distance between any pair of points in the space.
However, the Clifford torus is not trivially the same as the hypersphere. One salient aspect is that the Clifford torus maps directly and distortion-free back onto the flat square torus, whereas the hypersphere does not.

%%%%%%%%%%%%%%%%%%%%%%%%%%%%%%%%%%%%%%%%%%%%%%%
\subsection{Additional Training Considerations}

In order to ensure successful training in practice, some interactions between the representation geometry and the process of gradient descent must be noted.

Firstly, it may be advisable to ensure that training leads to a representation that efficiently uses the full bit range. This maximises the representational capacity of the chosen numeric precision, and also ensures that the geometry of the data takes on the geometry of the space.
For this purpose, we follow recent work in the use of ``KoLeo'' regularisation \citep{Sablayrolles:2018}. Inspired by the Kozachenko--Leonenko differential entropy estimator, this regulariser creates a repulsive force between near neighbour data points, promoting uniformity of data distribution
(Figure \ref{fig:torus_spreading}).

\begin{figure}[t]
  \centering
  \centerline{\includegraphics[width=0.4\columnwidth]{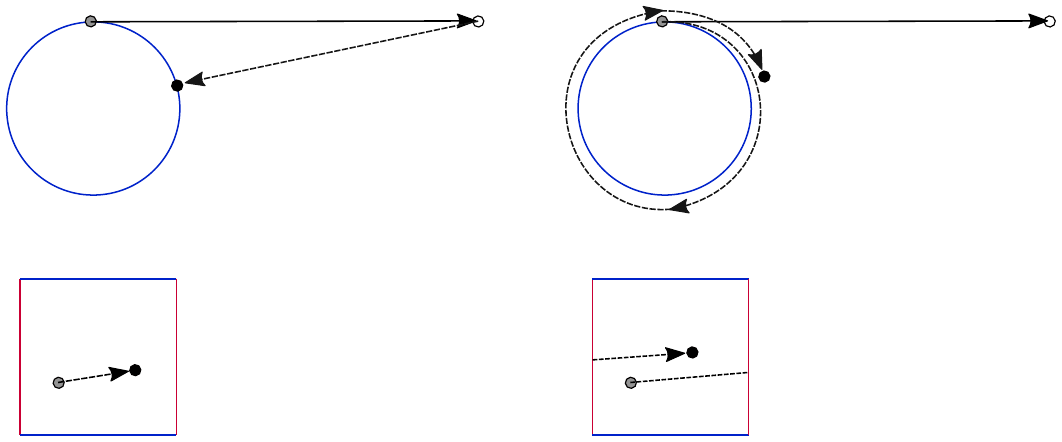}}
  \caption{Extremely large gradient updates have different effects in the Clifford torus, depending on whether produced by L2 normalisation or Clifford projection. This illustration shows a single update in a figurative Clifford torus (upper, circle) and a `view' of the same as a flat torus (lower, square). Under L2-normalisation (left), a gradient update is applied in the extrinsic space (solid arrow) before normalisation (dashed arrow), with the side-effect that gradient updates of unbounded size will have a stable impact---in fact, asymptotically limited to a 90 degree angle. In the Clifford-projected space (right), very large gradient updates simply `wrap around', which means that very small deviations in large steps (e.g.\ floating point rounding errors) may lead to unstable updates.}
  \label{fig:gradient_clipping_torus}
\end{figure}

Second, we find that gradient clipping is advisable, since in piloting we observed that some configurations exhibited a tendency to diverge.
In an L2-normalisation approach, large gradients are limited in their effect, because any very large gradient update in the unconstrained space is then projected back onto the hypersphere, with constrained effect (Figure \ref{fig:gradient_clipping_torus}, left). In fact, when L2-normalisation is included in the procedure, the outcome is that a gradient update step is replaced by an angular step whose magnitude follows the \textit{tan()} of the step size in the unconstrained space. This is a beneficial side-effect of the topological compromise used in L2-normalisation-based training, where gradient updates are applied in the extrinsic space and do not align perfectly with the intrinsic geometry.
Conversely, under the Clifford projection \eqref{eq:cliffordD} a very large gradient update may ``wrap around'' the space multiple times, leading to unstable parameter updates (Figure \ref{fig:gradient_clipping_torus}, right). Small floating-point differences in a large step size could lead to quite different outcomes when this happens. It is by far preferable that update steps should usually be smaller than the diameter of the space.

We also encountered a second motivation for gradient clipping. We have found during initial studies with CIFAR data that KoLeo regularisation can create instabilities at the beginning of training, especially when the dimensionality is low. The gradient of the KoLeo loss scales as the inverse of nearest-neighbour distances, and when these distances are very small there is a strong influence of $1/\epsilon$ on the gradient, which would be very large values. The KoLeo loss function could be modified to reduce this effect, but we consider it preferable to maintain the form of the loss (and thus the direction of the gradients) while clipping the gradient values.

As a further minor issue, we note that many of the projections considered here involve constant scaling factors such as $2 \pi$ or $\sqrt D$. These can often be omitted as code optimisations, since these fixed multipliers do not change the shape of the loss landscape.

%%%%%%%%%%%%%%%%%%%%%%%%%%%%%%%%%%%%%%%%%%%%%%%%%%%%%%%%%%%%%%%%%%%%%%%%%%%%%%%%%%%%%%%%%
%%%%%%%%%%%%%%%%%%%%%%%%%%%%%%%%%%%%%%%%%%%%%%%%%%%%%%%%%%%%%%%%%%%%%%%%%%%%%%%%%%%%%%%%%
%%%%%%%%%%%%%%%%%%%%%%%%%%%%%%%%%%%%%%%%%%%%%%%%%%%%%%%%%%%%%%%%%%%%%%%%%%%%%%%%%%%%%%%%%
\section{Inference in Torus Embeddings}

Inference in torus embeddings can use the Clifford space, i.e.\ the same unit-norm representations as we have discussed for training (indeed we use this in our main experiment).
However, a specific benefit of torus topology is that our representations have a 1-to-1 correspondence with the flat square hypertorus, which can enable distance queries to be implemented with an extremely small number of processor instructions.

Consider the case in which hypertoroidal data has already been calculated by some process, and we wish to perform efficient distance calculations such as for similarity search.
For any data represented in Clifford fashion, the conversion to flat torus representation is with the use of \textit{arctan2()}. The flat torus representation, further mapped on to the integers, is a desirable representation e.g.\ to create compact lookup tables.
This is the format we would advocate for general use.
High-entropy integers are a compact way to store information on disk and in RAM.

In such a representation, distance search can be extremely efficient: a one-dimensional shortest distance is calculated as $\min({a-b},{b-a})$, allowing the subtraction operations to overflow (wrap around), which will happen in many systems using uint8 or similar when the representation uses the full number range.
One-dimensional distances can then be accumulated in the usual fashion to find the $D$-dimensional $L_p$-distance.

Calculation efficiency at inference time is also affected by any preprocessing and lookup operations needed. if data have been compressed by a scheme such as PQ, for example, then a comparison requires a lookup operation, with a lookup table (LUT) in memory of size $k^* D$ floats (where $k^*$ is the number of PQ centroids per subspace) \citep{Jegou:2011}. The equivalent in a torus projection would simply be the linear scaling parameters to map the input into the space, i.e. $2 . D$, thus %in practice always
smaller in memory, with fewer operations needed, and lower implementation complexity.
Table \ref{tbl:quants} illustrates these tradeoffs for the configurations we use in our study.

%%%%%%%%%%%%%%%%%%%%%%%%%%%%%%%%%%%%%%%%%%%%%%%%%%%%%%%%%%%%%%%%%%%%%%%%%%%%%%%%%%%%%%%%%
%%%%%%%%%%%%%%%%%%%%%%%%%%%%%%%%%%%%%%%%%%%%%%%%%%%%%%%%%%%%%%%%%%%%%%%%%%%%%%%%%%%%%%%%%
%%%%%%%%%%%%%%%%%%%%%%%%%%%%%%%%%%%%%%%%%%%%%%%%%%%%%%%%%%%%%%%%%%%%%%%%%%%%%%%%%%%%%%%%%
\section{Experiments}

We explore the characteristics of torus embeddings through simple experiments using standard deep learning practices: in particular, contrastive learning with distance-based loss functions \citep{Khosla:2020%supcon
}.
We conducted three experimental tests:
\begin{enumerate}
\item We first trained networks with the CIFAR datasets for image classification \citep{Krizhevsky:2009}, exploring various aspects of training torus embeddings, in particular the two different torus mappings.
\item We then explored the effects of post-training quantisation on the performance of these networks, using CIFAR100.
\item We followed this with a study on few-shot classification of acoustic birdsong data, to evaluate the method in a different context, with a large audio dataset (`BIRB') designed to probe generalisation.
\end{enumerate}

Our Python implementation is available online, separately for the image experiments\footnote{\url{https://github.com/danstowell/SupCon-Framework/tree/torus}}
and the audio experiment.\footnote{\url{https://github.com/danstowell/ProtoCLR/tree/torus}}

\subsection*{Experiment 1: CIFAR training}

We first train via CIFAR image classification \citep{Krizhevsky:2009}.
We evaluate our learnt representations in the paradigm of information retrieval, using Precision-at-1 as our main measure of retrieval performance.
To train each representation we use a fixed budget of 100 epochs per run, with the Adam optimiser, gradient clipping, and early stopping.
For brevity, in the following we refer to torus embeddings as follows: \textit{torusC} for an embedding produced by the Clifford projection, and \textit{torusN} for an embedding produced by the L2p normalisation projection.

Experimental runs that used the torusC projection did not always converge: some diverged, and some failed to learn. This came despite the use of gradient clipping: although gradient clipping helped to stabilise torusC, the gradient clipping would have needed to be unhelpfully strong to ensure convergence in all cases.
In our experiments the addition of KoLeo regularisation further helped to prevent divergence in torusC. However, strong KoLeo regularisation at low dimension sometimes had a counterproductive effect of its own: extremely large gradients often occurred due to the influence of very close near neighbours, which could lead to divergence. This can also be ameliorated by gradient clipping.
Having observed these effects in pilot runs, we then chose a fixed gradient clipping threshold (100) to use throughout the main experiment, and we evaluated multiple KoLeo weights.

When comparing hypertorus against hypersphere training runs, we maintained the same dimensionality in the pre-projection representation, for maximum comparability.
As noted, the embedding projections variously affect the extrinsic and intrinsic dimensionality, which we keep in mind when comparing outcomes.

We evaluated embeddings primarily via retrieval performance.
We also explored whether our training is effective at spreading data points throughout the representation space.
As mentioned, KoLeo regularisation is intended to help with this; the geometry and dimensionality of the space may also have a significant influence.
To measure the spread of data we choose the \textit{circular variance} common in directional statistics. Given $Z$, a set of $n$ vectors of unit L2 norm, the circular variance is calculated as
\begin{equation}\label{eq:circvar}
1 - \left|\left|\frac{1}{n}\sum_{i=1}^n{Z_i}\right|\right|_2
\end{equation}
To evaluate the circular variance \eqref{eq:circvar} for a given representation, we calculate it across all the CIFAR data (training and validation) in order to give the best estimate of the resultant data spread.
Since the statistic is calculated on unit-norm vectors, for torus representations we use the data in their Clifford representation.

\subsection*{Experiment 2: quantisation}

After evaluating the floating-point representations, we also apply the same analysis to discrete representations produced by post-training quantisation.
The goal of this is to explore how well toroidal representations perform under quantisation and compression, and how this compares against their equivalent hyperspherical representations.
We evaluate a simple `n-bit' grid quantisation, as well as a more advanced product quantisation (PQ) method.

Grid quantisation is selected for simple implementation with minimal code and memory requirements. Our aim with n-bit grid quantisation is to select a quantisation that maps directly onto the fundamental n-bit integer representation (with overflow) in processors.
Hence we investigate the viability of using grid-quantised values directly, using 8 bits or 1 bit per vector.
Note that 1-bit quantisation is an extreme case, in which all $L_p$ distance measures become equivalent to Hamming distance.
Although binary vector spaces can be interpreted as having hypertorus structure, toroidal distance measures would not be distinct from others.
Thus the only role of 1-bit quantisation in the present work is to explore extreme compression of our trained representations.

As an alternative to grid quantisation, for modern high-compression quantisation we select product quantisation (PQ), implemented in the `faiss' software library. It achieves a very high compression rate for high-dimensional vectors, with efficient querying \citep{Jegou:2011}.
PQ requires training (which we perform after the main training phase, and apply to the 8-bit quantised values). Once trained, searching or decoding is efficient to implement, given the lookup table stored in memory (whose size can be configured at training time).
A toroidal space is a product space, and so in principle should be a good match for product quantisation.

\begin{figure}[t]
  \centering
  \includegraphics[width=\columnwidth,page=1]{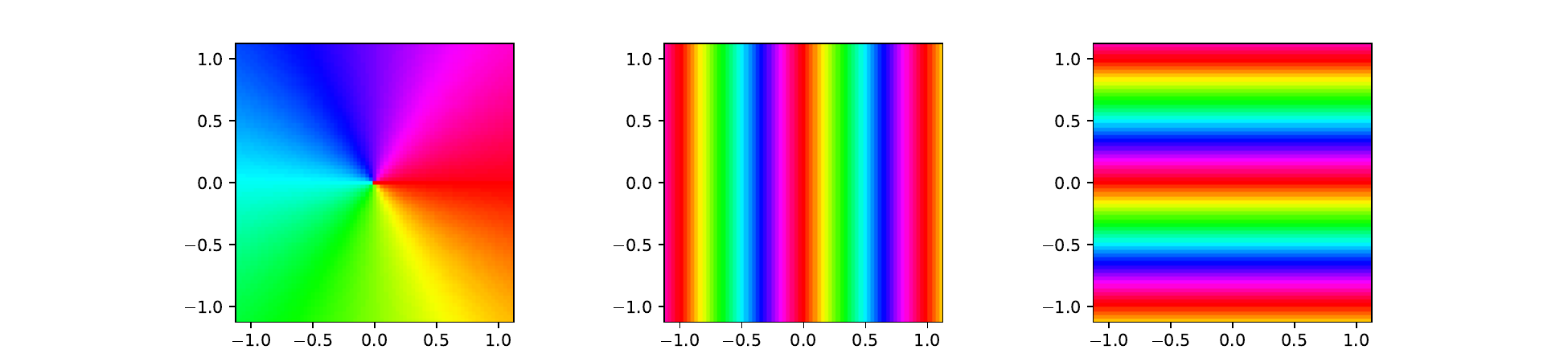}
  \caption{The two methods for mapping data onto a hypertorus use very different mappings, as in these 2D illustrations. Left panel: $L_{2p}$ normalisation maps 2D points to their angle with respect to the origin. Centre and right panels: the Clifford projection transforms each dimension separately into a cyclical version of itself.}
  \label{fig:rainbowquant}
\end{figure}

Torus embeddings offer us a further choice regarding quantisation, that is not available in other geometries: we can quantise either in the flat torus space or in the Clifford representation. For grid quantisation, this amounts to quantising on a grid of either the scalar values or the angles, with quite different implications (cf.\ Figure \ref{fig:rainbowquant}).
The geometry of the flat torus is preferred since the data can `fill' this space directly in its extrinsic representation, and it is more likely that the axes of this flat geometry can be treated independently for quantisation.
Thus for 8-bit quantisation and for PQ, we use the flat torus representation.
The traditional hypersphere cannot be mapped into a flat geometry without distortion, so we apply quantisation to the embedded hypersphere directly.

Regarding the bit-rate of the resulting encodings:
n-bit grid encoding results in $n . D$ bits for a vector.
To quantise the torusN embedding, we convert it from a Clifford space to a flat torus (using arctan2) which halves the dimensionality, and thus the total bits-per-vector in the torusN case is $n . D / 2$.

The resulting set of quantisations covers a range of codebook sizes and compressed vector sizes, focussed on low bitrates
(Table \ref{tbl:quants}).
The lowest rates use 8 bits per vector---i.e.\ only 256 possible values, an extreme case for a 100-class classification task (i.e.\ CIFAR100).

\begin{table}[t]
\centering
\begin{tabular}{c|rrr}
Quantisation method & Codebook size &  Bits per vector & Num possible values\\
\hline
8-bit    & $  2D$ & $8D$ *            & $256^D$ \\
1-bit    & $  2D$ & $D$ \hspace{1.5mm}  & $2^D$ \\
PQ(8,16) & $256D$ & 128 \hspace{1.5mm}  &  $3.4 \times 10^{38}$ \\
PQ(8,4)  & $256D$ &  32 \hspace{1.5mm}  &  $4.2 \times 10^9$   \\
PQ(8,2)  & $256D$ &  16 \hspace{1.5mm}  & 65536 \\
PQ(8,1)  & $256D$ &   8 \hspace{1.5mm}  &   256 \\
PQ(4,4)  & $ 16D$ &  16 \hspace{1.5mm}  & 65536 \\
PQ(4,2)  & $ 16D$ &   8 \hspace{1.5mm}  &   256 \\
\end{tabular}
  \caption{Quantisation methods applied. PQ(m,n) represents product quantisation with $m$ subspaces and $n$ bits used for indexing within each subspace. $D$ represents the extrinsic dimension of the input data. %
Codebook size is the number of full-precision scalar values needed in memory.
*Note that for 8-bit quantisation in the case of method torusN, the bits-per-vector is half that implied here, because we convert data from Clifford into the flat torus representation before quantisation, which halves the extrinsic dimension $D$.}
  \label{tbl:quants}
\end{table}

\subsection*{Experiment 3: prototypical learning with birdsong audio}

To explore whether the same characteristics of toroidal embedding spaces are evident in a different setting, we performed a separate experiment on few-shot classification of acoustic birdsong data, using the prototypical contrastive learning (\textit{ProtoCLR}) method of \citet{Moummad:2024}. We also used their software framework, adding torusN as a representation mode and comparing it against the default hyperspherical representation.
Since this experiment uses more data and takes longer for each run, we did not evaluate all the aforementioned options, rather the core comparison of hypersphere versus torusN.

In common with the CIFAR experiments, here we used supervised training with a distance-based contrastive loss function, but this time for audio data.
The audio data are converted to spectrograms, as is common in the field, and the task is multi-class classification of bird species.
For this experiment we use the BIRB benchmark, which is designed to evaluate generalisation in bird sound recognition \citep{Hamer:2023}.
The training data come from the crowdsourced Xeno-canto library \citep{Vellinga:2015}.
The test data comprise six separate birdsong datasets, each representing different recording conditions and localities, as well as different sets of birds to classify.
Whereas the CIFAR datasets have 50,000 items for training and 10,000 for test,
BIRB has 684,000 items in its core training set, and six test sets of between 7,000 and 60,000 items.

Further, instead of ordinary supervised classification, in this experiment the evaluation is by few-shot classification.
We evaluate the trained systems for their accuracy in 1-shot and 5-shot scenarios, following \citet{Moummad:2024}.
For more detail on the experimental setup we refer the reader to that work.
Specific configuration choices we highlight are:
Each few-shot performance metric is calculated using ten times using different seeds, reporting the mean accuracy to generalise over the variability due to few-shot sampling.
KoLeo regularisation is not implemented in this case, in particular since we did not pursue the torusC variant.
The optimiser was stochastic gradient descent.
We do not perform hyperparameter tuning, but use the same hyperparameters selected in prior work across all runs.
The reported outcomes are thus internally comparable, even though higher performance may be possible.

%%%%%%%%%%%%%%%%%%%%%%%%%%%%%%%%%%%%%%%%%%%%%%%%%%%%%%%%%%%%%%%%%%%%%%%%%%%%%%%%%%%%%%%%%%%%%%%%%%%%%%%%%%%%%%%%%%%%%%%%
\subsection*{Results using CIFAR}

\begin{figure}[t]
  \centering
\resizebox{\linewidth}{!}{
  \includegraphics[width=0.41\columnwidth,page=1,clip,trim=0mm 0mm 25mm 0mm]{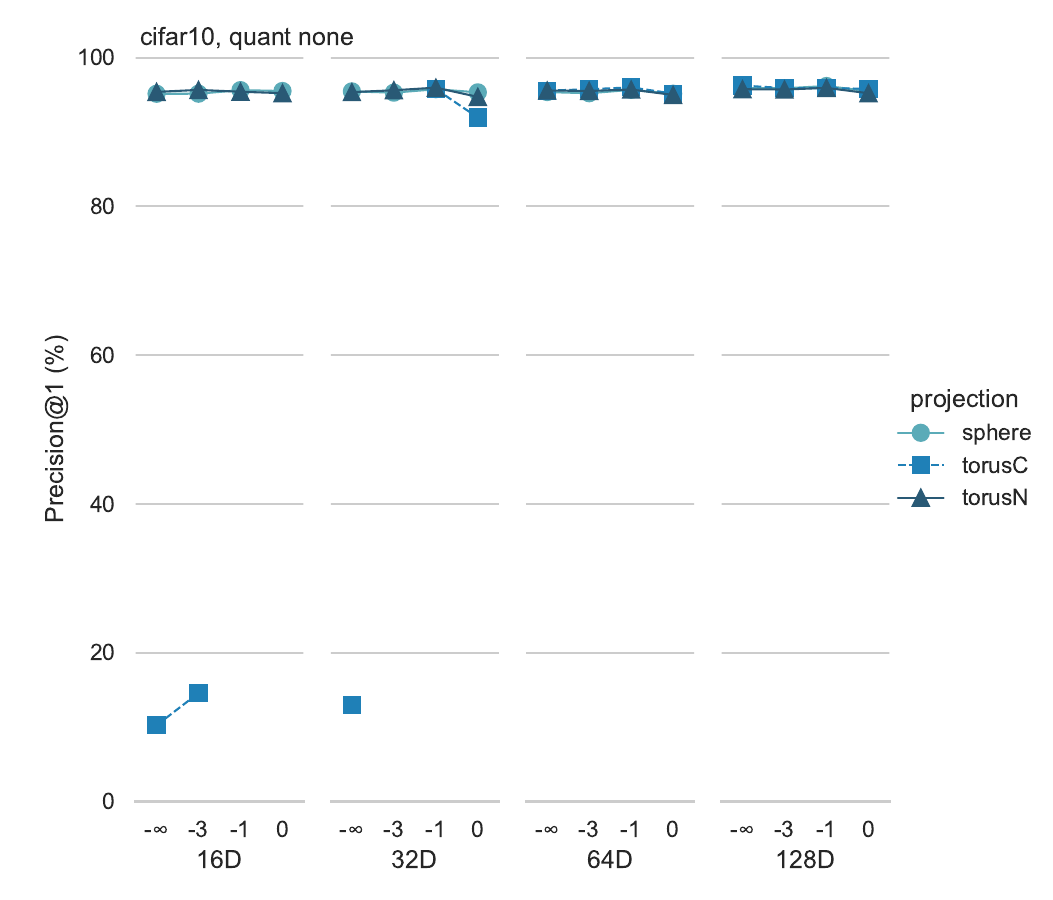}
  \includegraphics[width=0.427\columnwidth,page=8,clip,trim=19mm 0mm 0mm 0mm]{plot_evalmeasures_precision_at_1}
}
  \caption{Classification performance (precision at 1) on the CIFAR10 and CIFAR100 data sets, using nearest-neighbour classification, comparing embedding spaces that are hyperspherical versus hypertoroidal. The x-axis shows variation in the embedding dimensionality, and (nested inside) variation in the strength of the KoLeo regularisation (the number is the $\log_{10}$ of the regularisation strength).}
  \label{fig:plot_evalmeasures_precision_at_1_raw}
\end{figure}

We trained our network using SupCon on CIFAR10 and CIFAR100, varying the nature of the embedding space: hyperspherical, torusC, or torusN, each at various dimensionalities (Figure \ref{fig:plot_evalmeasures_precision_at_1_raw}).
Most configurations performed equally well, and very strongly, on CIFAR10. We thus focus our discussion on CIFAR100.
The hyperspherical representation generally performed most strongly, with torusN producing competitive results. The torusC mapping proved to be unstable at low dimensionality. The first general finding is that torusN is strongly preferred over torusC for stability and good performance, especially at lower dimensionality.

Koleo regularisation had a dimensionality-dependent effect on the results: there appears to be a ``sweet spot'' for the Koleo weight, and this varies across method and dimensionality.
The Koleo regularisation appears to have a slightly stronger influence on the performance of torus representations than the equivalent hyperspherical representation.
The reasons for this are unclear, though potentially influenced by the difference in the intrinsic dimensionality.

\begin{figure}[t]
  \centering
\resizebox{\linewidth}{!}{
  \includegraphics[width=0.31\columnwidth,page=9, clip,trim=0mm 0mm 25mm 0mm]{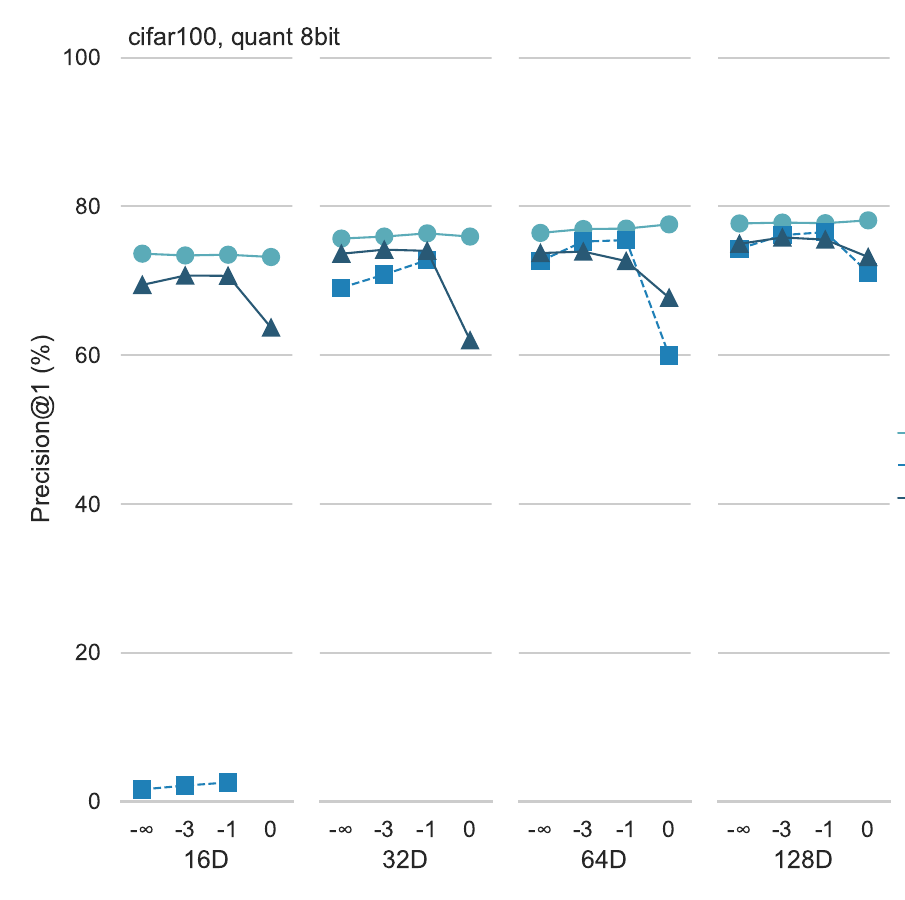}
  \includegraphics[width=0.27\columnwidth,page=10,clip,trim=19mm 0mm 25mm 0mm]{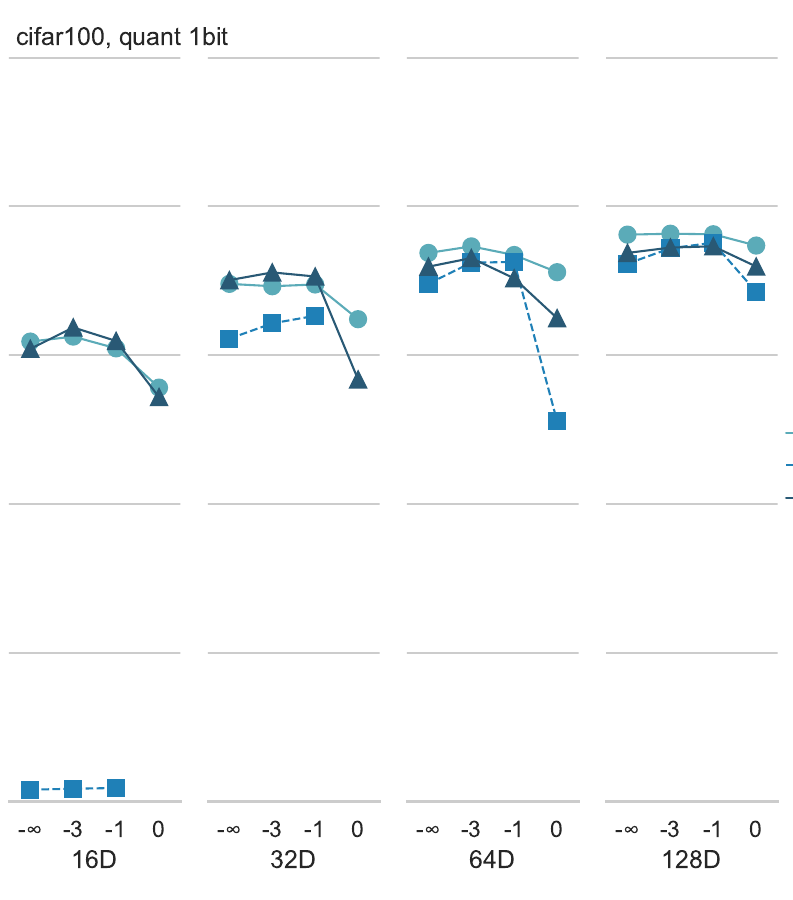}
  \includegraphics[width=0.32\columnwidth,page=11,clip,trim=19mm 0mm 0mm 0mm]{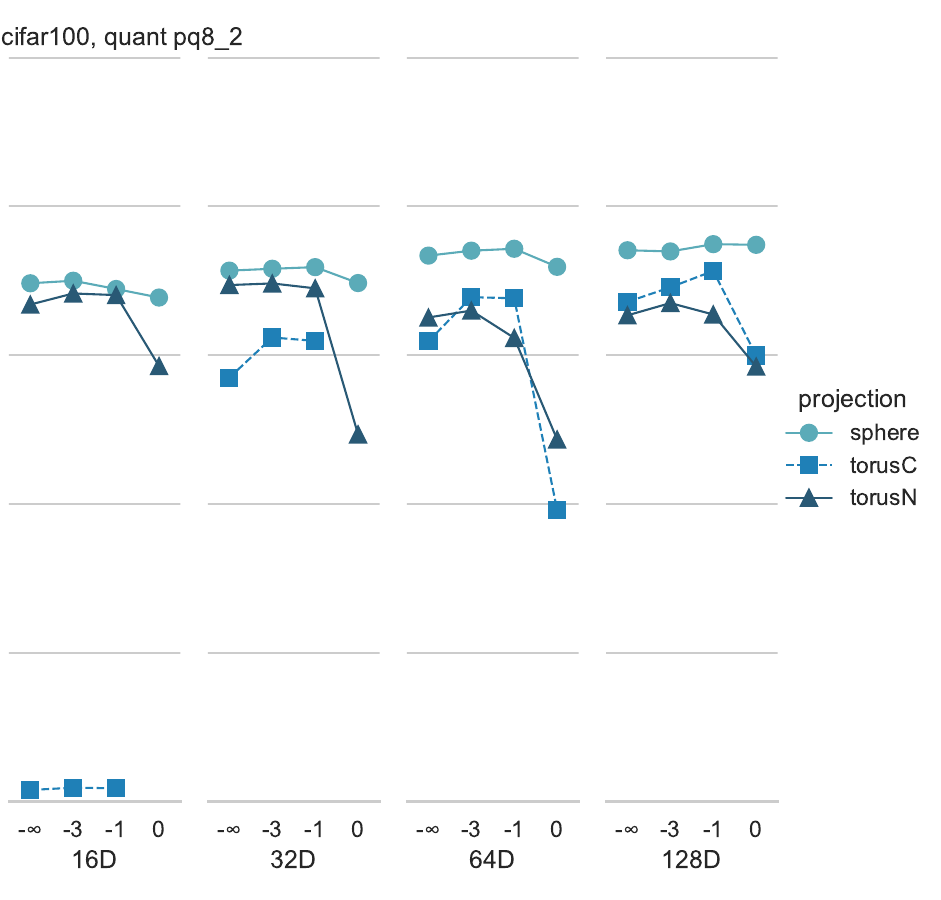}
}
\resizebox{\linewidth}{!}{
  \includegraphics[width=0.31\columnwidth,page=12,clip,trim=0mm 0mm 25mm 0mm]{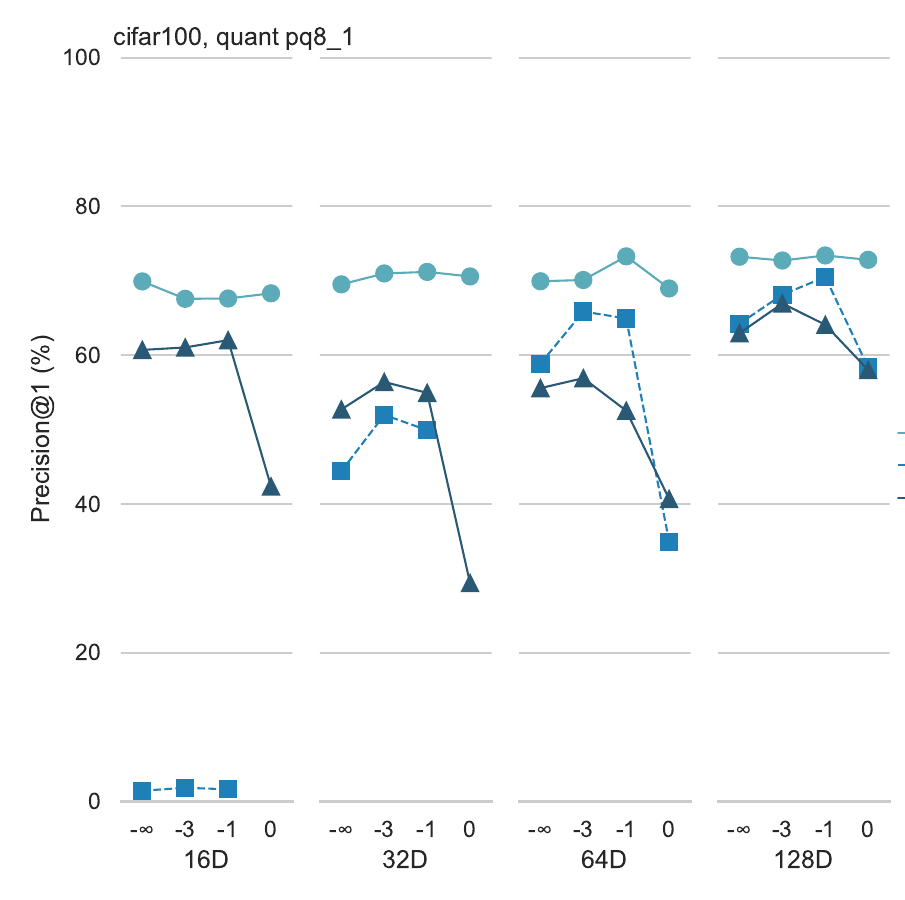}
  \includegraphics[width=0.27\columnwidth,page=13,clip,trim=19mm 0mm 25mm 0mm]{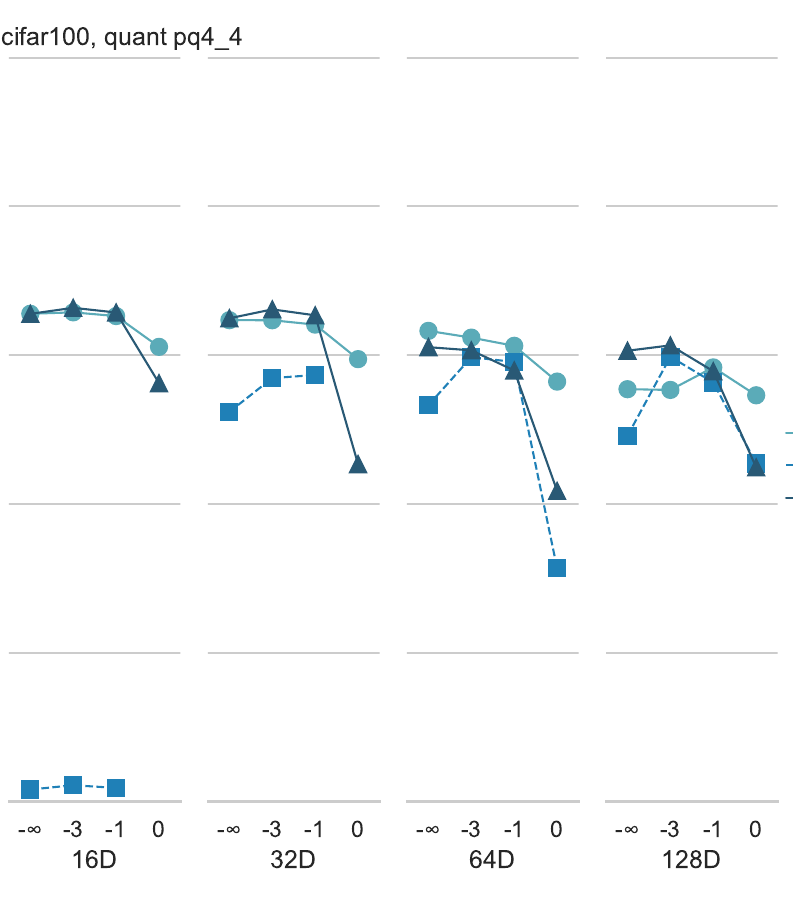}
  \includegraphics[width=0.32\columnwidth,page=14,clip,trim=19mm 0mm 0mm 0mm]{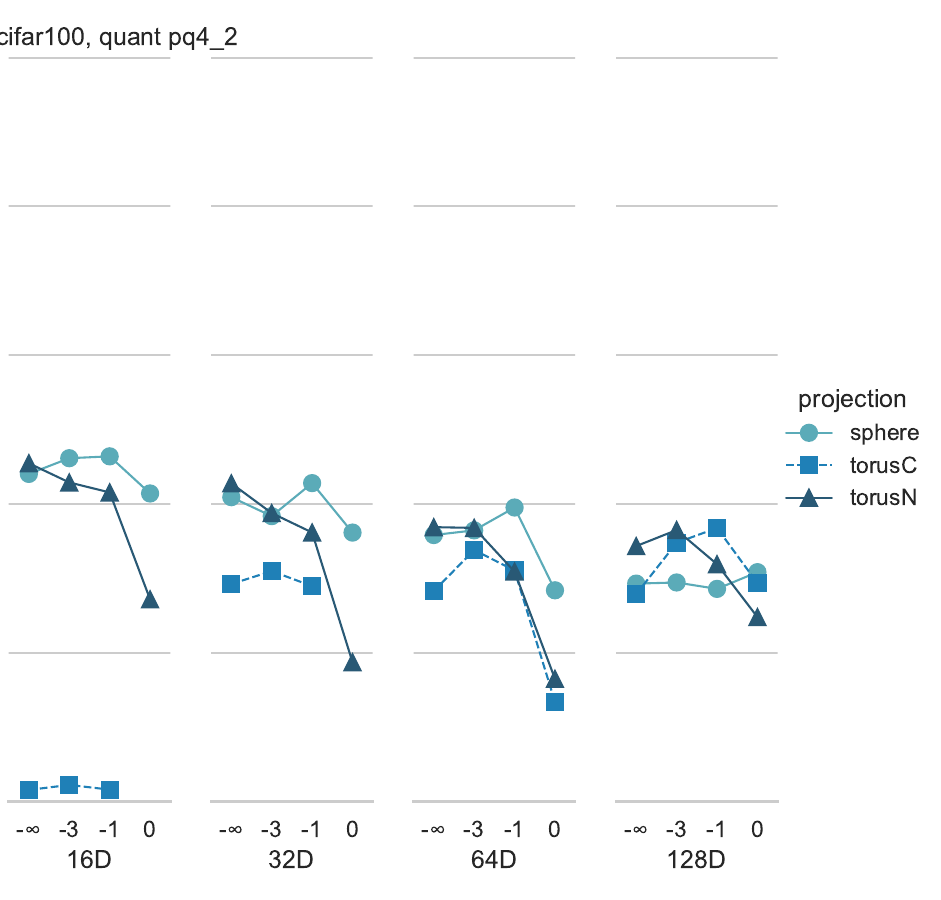}
}
  \caption{Classification performance as in Figure \ref{fig:plot_evalmeasures_precision_at_1_raw} (CIFAR100 data set), after quantisation by various methods. The 8-bit quantisation has only a mild impact on performance, but other methods degrade the classifier precision to varying extents.}
  \label{fig:plot_evalmeasures_precision_at_1_quant}
\end{figure}

Concerning the spread of vectors in the representation spaces, the circular variance shows a consistent pattern across hyperspherical and toroidal spaces, with KoLeo weight having a strong influence
(Appendix, Figure \ref{fig:torus_spreading} and Figure \ref{fig:plot_evalmeasures_circvar}).
At the strongest KoLeo weight, data points are very evenly spread through the space---although this objective competes with classification performance.

\subsection*{Results after quantisation}

The representations created in this study generally cope very well with quantisation (Figure \ref{fig:plot_evalmeasures_precision_at_1_quant}).
8-bit quantisation has very little effect on the performance on CIFAR100, as judged against the floating-point representations.
Taking this to an extreme, 1-bit quantisation generally applies sufficient compression to degrade the classification performance, as do the high-compression PQ settings take that even further.
In low data-rate and low dimension settings, torusN often outperforms hyprs, though not always.

While retrieval in 8-bit quantised space performed very well, the data compression ratio is not dramatic, since grid quantisation still requires $n.D$ bits for each vector.
Using PQ we compressed to even lower bit-rates, often even lower than the 1-bit grid. As has been observed by others, PQ provides a representation that is very resilient for low bit-rate.
Compressing to 8 or 16 bits per vector, but allowing a large codebook size, PQ(8,2) and PQ(8,1) yielded very good performance for hyperspherical representations.
Imposing stronger restrictions on the codebook size, PQ(4,4) and PQ(4,2) incurred performance penalties across all configurations. In these extreme conditions, both hyperspherical and torusN performed equally well, outperforming each other equally often.
We note also that these extreme conditions favoured a low-dimensional representation as input, better than a high-dimensional one. This is the reverse of the results on the uncompressed representations, and indicates that there may be a bottleneck interaction between the data representation and its downstream quantisation.

Despite our expectations, the quantised hypersphere did not underperform the quantised torus in general.
This is notable given that PQ is based on a high-dimensional `grid' of centroids, as a side-effect of its factorisation of the data space.
A representation such as the flat square torus should be well adapted to grid quantisation, due to its rectangular shape, while a hypersphere is not intrinsically a good match.
Despite this, hyperspherical representations can be quantised well by PQ even at high compression.

\subsection*{Results of prototypical learning with birdsong audio}

\begin{figure}[t]
\centering
\resizebox{\linewidth}{!}{
\includegraphics[width=0.83\columnwidth,page=1,clip,trim=0mm 10mm 0mm 0mm]{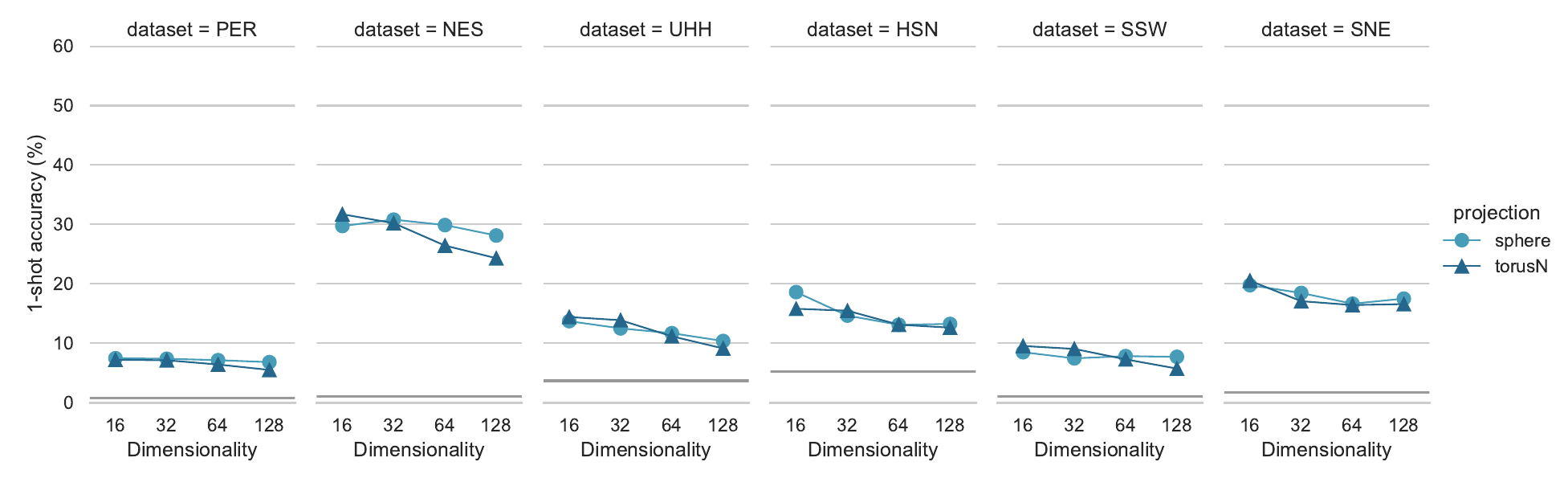}%
\includegraphics[width=0.12\columnwidth,page=3,clip,trim=23mm 10mm 28mm 3mm]{plot_evalmeasures_protoclr}
}
\resizebox{\linewidth}{!}{
\includegraphics[width=0.83\columnwidth,page=2,clip,trim=0mm 0mm 0mm 11mm]{plot_evalmeasures_protoclr}%
\includegraphics[width=0.12\columnwidth,page=4,clip,trim=23mm 0mm 28mm 10mm]{plot_evalmeasures_protoclr}
}
\caption{Birdsong audio classification performance (accuracy) in a few-shot setting, comparing embedding spaces that are hyperspherical versus hypertoroidal. Scores shown are for each of six birdsong datasets provided by BIRB (left to right), with the overall average summarised in the rightmost panel. A dark gray line indicates chance performance for each dataset. The x-axis shows variation in the embedding dimensionality. Upper: 1-shot. Lower: 5-shot. }
  \label{fig:plot_evalmeasures_protoCLR}
\end{figure}

Using the BIRB benchmark of bird audio data, with prototypical contrastive learning, both hyperspeherical and toroidal representations trained well and achieved similar levels of performance
(Figure \ref{fig:plot_evalmeasures_protoCLR}).
There was large variation in performance between the six test datasets, as expected due to their varying complexities \citep{Hamer:2023}.
We saw a consistent pattern of lower-dimensional representations yielding better generalisation performance in this few-shot setting.
Averaging across the datasets, the hypersphere representation gave better performance in 128D ($+1.6\%$/$+2.7\%$ for 1-shot/5-shot respectively), there was little difference in 64D, and the torus representation gave better performance for 32D (($+0.3\%$/$+2.9\%$) and 16D ($+0.2\%$/$+2.5\%$).
The overall best configuration in this test was the 16D torus, for both 1- and 5-shot retrieval.

%%%%%%%%%%%%%%%%%%%%%%%%%%%%%%%%%%%%%%%%%%%%%%%%%%%%%%%%%%%%%%%%%%%%%%%%%%%%%%%%%%%%%%%%%
%%%%%%%%%%%%%%%%%%%%%%%%%%%%%%%%%%%%%%%%%%%%%%%%%%%%%%%%%%%%%%%%%%%%%%%%%%%%%%%%%%%%%%%%%
%%%%%%%%%%%%%%%%%%%%%%%%%%%%%%%%%%%%%%%%%%%%%%%%%%%%%%%%%%%%%%%%%%%%%%%%%%%%%%%%%%%%%%%%%
\section{Discussion}

The goal of this investigation was to explore torus embeddings---a variation in method inspired by hyperspherical embeddings---and evaluate them within a DL framework.

A potential benefit is the direct mapping to a ``flat square'' space that quantises trivially to ordinary integer representations. Potential drawbacks are the reduced intrinsic dimensionality (for torusN) or instabilities in training (for torusC). We established that the torusN method achieves performance results very similar to a hyperspherical representation, and trains well, with no apparent inhibition in quality despite the lower intrinsic dimension.
On both images and audio data, the L2p-normalised torus representation \eqref{eq:L2p} led to performance similar to that of hyperspherical embeddings and with potential benefits at low dimension and high compression.
This was verified in a challenging generalisation benchmark for birdsong audio.

We note that the performance differences between hyperspherical and toroidal representations are small compared to the impacts of other configuration choices.
For consistent performance on a small dataset such as CIFAR, a hypertorus model benefits from KoLeo regularisation (and gradient clipping for torusC) to ensure training progresses smoothly. These observations also provide insight into the functioning of the popular hyperspherical embedding: L2 normalisation is a projection which inherently mitigates the impact of extreme gradient updates, by limiting the possible step size. This benefit carries over to one of our torus projections but not the other.

An additional hypothesis was that the flat torus space being well-matched to lattice grid quantisation, it might show especially robust performance after quantisation. This was not observed: a well-trained model in either toroidal or spherical space performed very well under lattice quantisation, including extreme 1-bit quantisation. Further, product quantisation showed very strong performance in all our model configurations, despite the overhead of the added complexity in implementation.

We have investigated 1-bit quantisation of our vector dimensions. 1-bit representations can also be the output from binary neural networks (BNNs), which have been widely investigated
\citep{Rastegari:2016}%xnor-net
\citep{Menghani:2023}%survey
. To our knowledge, however, it has rarely been remarked that binary representations can be interpreted as having toroidal geometry: Integer ``modulo 2'' data can take only values 0 and 1. In modulo 2, adding two numbers has the same result as binary XOR, while multiplying them has the same result as binary AND (these are operations commonly used to build BNNs).
We do not consider that toroidal geometry should be used directly within BNNs. However, if torus and binary representations have matching topologies, this may be fruitful in future e.g.\ to distil toroidal representations onto BNNs.

%%%%%%%%%%%%%%%%%%%%%%%%%%%%%%%%%%%%%%%%%%%%%%%%%%%%%%%%%%%%%%%%%%%%%%%%%%%%%%%%%%%%%%%%%
%%%%%%%%%%%%%%%%%%%%%%%%%%%%%%%%%%%%%%%%%%%%%%%%%%%%%%%%%%%%%%%%%%%%%%%%%%%%%%%%%%%%%%%%%
%%%%%%%%%%%%%%%%%%%%%%%%%%%%%%%%%%%%%%%%%%%%%%%%%%%%%%%%%%%%%%%%%%%%%%%%%%%%%%%%%%%%%%%%%
\section{Conclusions}

The popularity of foundation models suggests a paradigm for the future of applied AI which is different from that previously envisaged: a small number of extremely large deep learning algorithms might be trained; algorithms for AI applications might then be derived from these, with further optimisation (such as fine-tuning or distillation) or even without (e.g. through few-shot learning, or in-context learning).
	The carbon footprint of this paradigm is a matter of choice and regulation: the prohibitively high environmental impact of (pre)training a massive model may be offset by the fact that very, very few such training runs might be needed by society.
The priority, then, is to ensure that these representations are highly suited to low-power distillation that can be used across common processors.

We have demonstrated that hypertorus representations can be trained in much the same way as hyperspherical representations.
Of the two methods we describe, the normalisation-based torus mapping yields the best results in terms of both stability and performance, while being very similar to the standard process used for hypersphere mapping.
Toroidal representations have a specific advantage: they are a very suitable match for common CPU integer representations.
We have demonstrated that they can produce quantised representations which perform well even at low dimensionality.

%%%%%%%%%%%%%%%%%%%%%%%%%%%%%%%%%%%%%%%%%%%%%%%%%%%%%%%%%%%%%%%%%%%%%%%%%%%%%%%%%%%%%%%%%
%%%%%%%%%%%%%%%%%%%%%%%%%%%%%%%%%%%%%%%%%%%%%%%%%%%%%%%%%%%%%%%%%%%%%%%%%%%%%%%%%%%%%%%%%
%%%%%%%%%%%%%%%%%%%%%%%%%%%%%%%%%%%%%%%%%%%%%%%%%%%%%%%%%%%%%%%%%%%%%%%%%%%%%%%%%%%%%%%%%
\acks{DS acknowledges support from the MSCA grant \textit{Bioacoustic AI} (101116715) and Horizon Europe grant \textit{MAMBO} (101060639).}

\bibliography{refs.bib}

%%%%%%%%%%%%%%%%%%%%%%%%%%%%%%%%%%%%%%%%%%%%%%%%%%%%%%%%%%%%%%%%%%%%%%%%%%%%%%%%%%%%%%%%%
%%%%%%%%%%%%%%%%%%%%%%%%%%%%%%%%%%%%%%%%%%%%%%%%%%%%%%%%%%%%%%%%%%%%%%%%%%%%%%%%%%%%%%%%%
%%%%%%%%%%%%%%%%%%%%%%%%%%%%%%%%%%%%%%%%%%%%%%%%%%%%%%%%%%%%%%%%%%%%%%%%%%%%%%%%%%%%%%%%%
\clearpage
\begin{appendices}

\section{Distances in hyperspheres and hypertori}

We illustrate some properties of the cosine distance in torus representations, compared against $L_p$ distances.

\begin{figure}[hpt]
  \centering
  \centerline{\includegraphics[width=0.8\columnwidth]{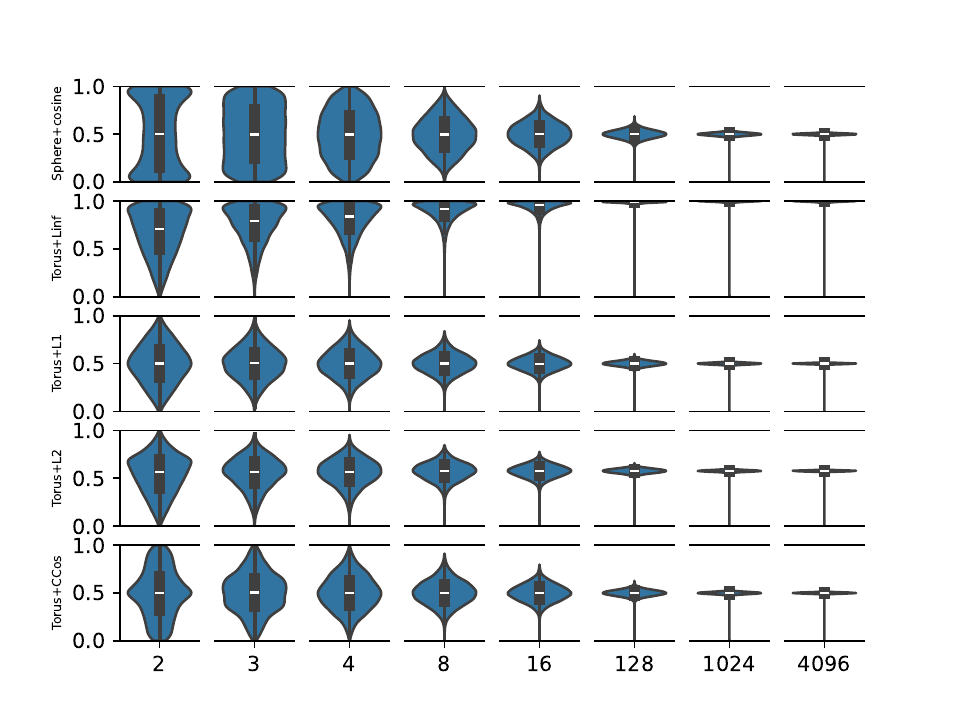}}
  \caption{Simulation to illustrate the effect of the curse of dimensionality in hyperspherical and hypertoroidal spaces. Distances are normalised against the largest possible in the space (the diameter of the space).}
  \label{fig:plot_torusdist_distribs}
\end{figure}

First, the effect of dimension, and the curse of dimensionality
(Figure \ref{fig:plot_torusdist_distribs}).
For each plot, 10,000 uniformly-distributed pairs of points were sampled in the appropriate space, and the distribtion of their distances calculated. From left to right we increase the dimensionality; from top to bottom we vary the geometry of the space and the choice of distance measure. In toroidal spaces, the L1 and L2 distances (calculated in the flat torus) or the cosine distance (calculated in the Clifford torus) each exhibit similar behaviour as the cosine distance applied to hyperspherical representations.

%%%%%%%%%%%%%%%%%%%%%%%%%%%%%%%%
\label{cosinelikeLp}

Next we consider the relation between cosine distance in a Clifford torus, and norms in the corresponding flat torus.

Consider the norm for a point $\textbf{a} = (a_0, a_1, ...)$ whose coordinates are given in the flat torus space, relative to an origin. Combining the Clifford projection with the cosine distance calculation, the distance of $\textbf{a}$ from the origin is:

$$|\textbf{a}|_\text{cc} = D - (\cos 2 \pi a_0 + \cos 2 \pi a_1 + ...)$$

\begin{figure}[hpt]
  \centering
  \centerline{\includegraphics[width=0.99\columnwidth]{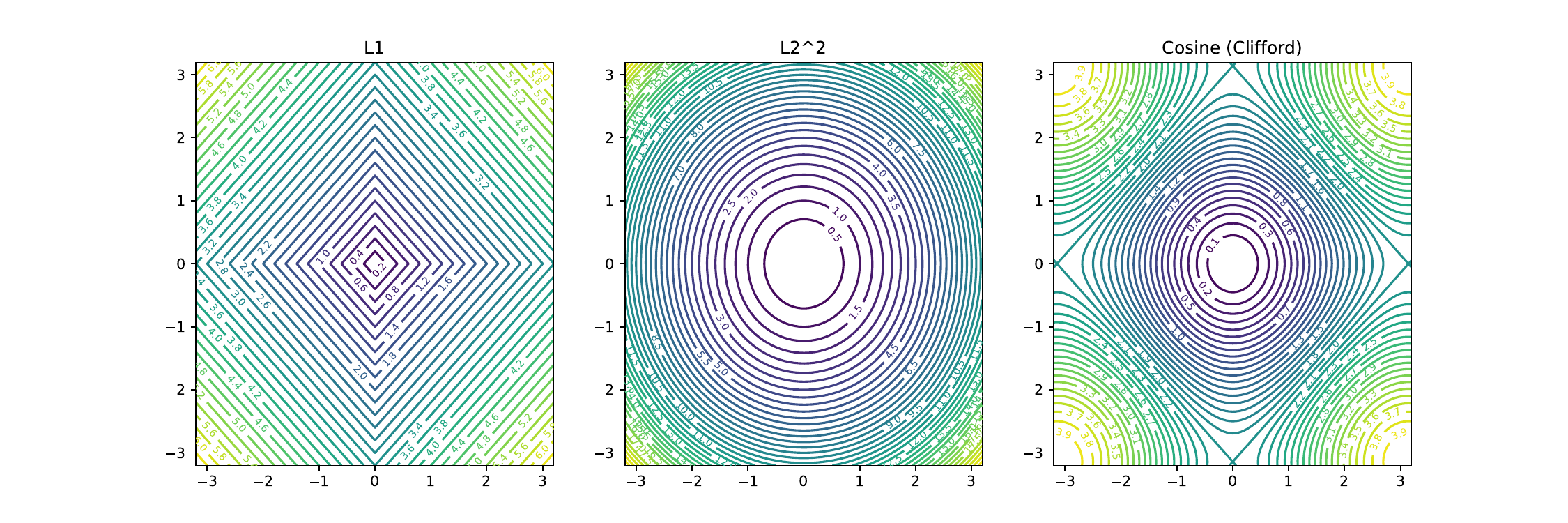}}
  \caption{Contour plot illustrating the characteristics of distance measures in a 2-dimensional torus. We focus on the quadrant closest to the origin: in the full torus all distance measures `reflect' at the boundaries shown here.}
  \label{fig:diamonds_contourplot}
\end{figure}

Taylor expansion of this function illustrates the behaviour, also visible in Figure \ref{fig:diamonds_contourplot}.
The Taylor expansion around the origin:
$$
\sum_{d=0}^{D-1}{
    \left(
        \frac{a_d^2}{2!} - \frac{a_d^4}{4!} + \frac{a_d^6}{6!} - ...
    \right)}
$$
is dominated by a quadratic term, which implies the distance behaves like $L_2^2$ at small values.
The Taylor expansion around the midpoint (any point where the dot products sum to zero so that $|\textbf{a}|_\text{cc}=D$) is:
$$
\sum_{d=0}^{D-1}{
    \left(
        a_d - \frac{a_d^3}{3!} + \frac{a_d^5}{5!} - ...
    \right)}
$$
and is dominated by a linear term, so at this point it behaves more like $L_1$.

%%%%%%%%%%%%%%%%%%%%%%%%%%%%%%%%
\clearpage
\section{Spreading vectors: KoLeo regularisation and circular variance}

\begin{figure}[hpt]
  \centering
  \includegraphics[width=0.24\columnwidth]{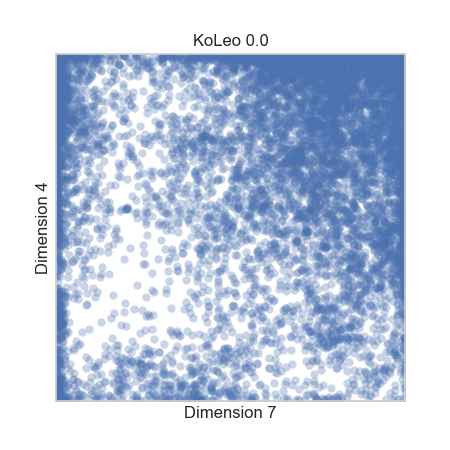}
  \includegraphics[width=0.24\columnwidth]{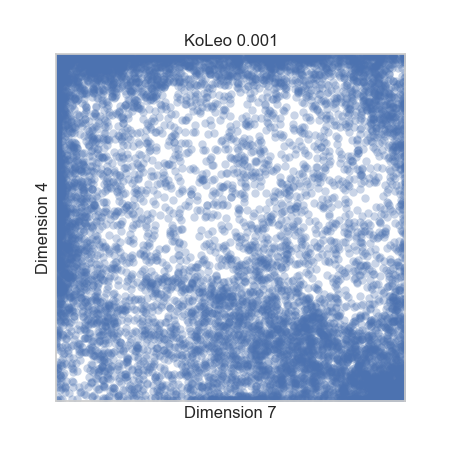}
  \includegraphics[width=0.24\columnwidth]{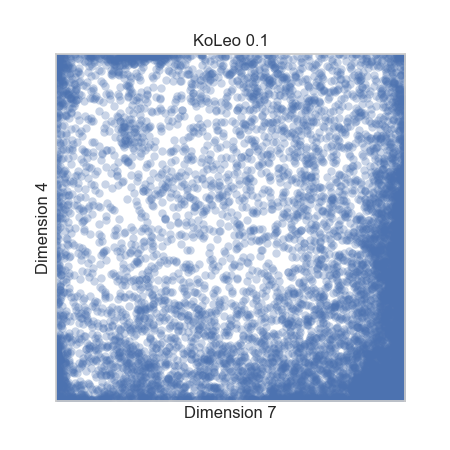}
  \includegraphics[width=0.24\columnwidth]{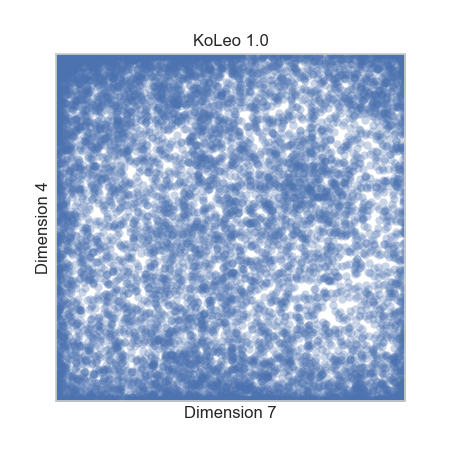}
  \caption{Effect of KoLeo regularisation. Here we visualise two arbitrary dimensions of the CIFAR100 validation set in a 128D trained hypertorus embedding (torusN), at various values of the KoLeo regularisation strength (cf.\ \citep{Sablayrolles:2018}). At the highest regularisation strength, data are spread more uniformly through the space, as is also reflected by the circular variance measurements below.}
  \label{fig:torus_spreading}
\end{figure}

\begin{figure}[hpt]
  \centering
  \includegraphics[width=0.31\columnwidth,page=8,clip,trim=0mm 0mm 0mm 0mm]{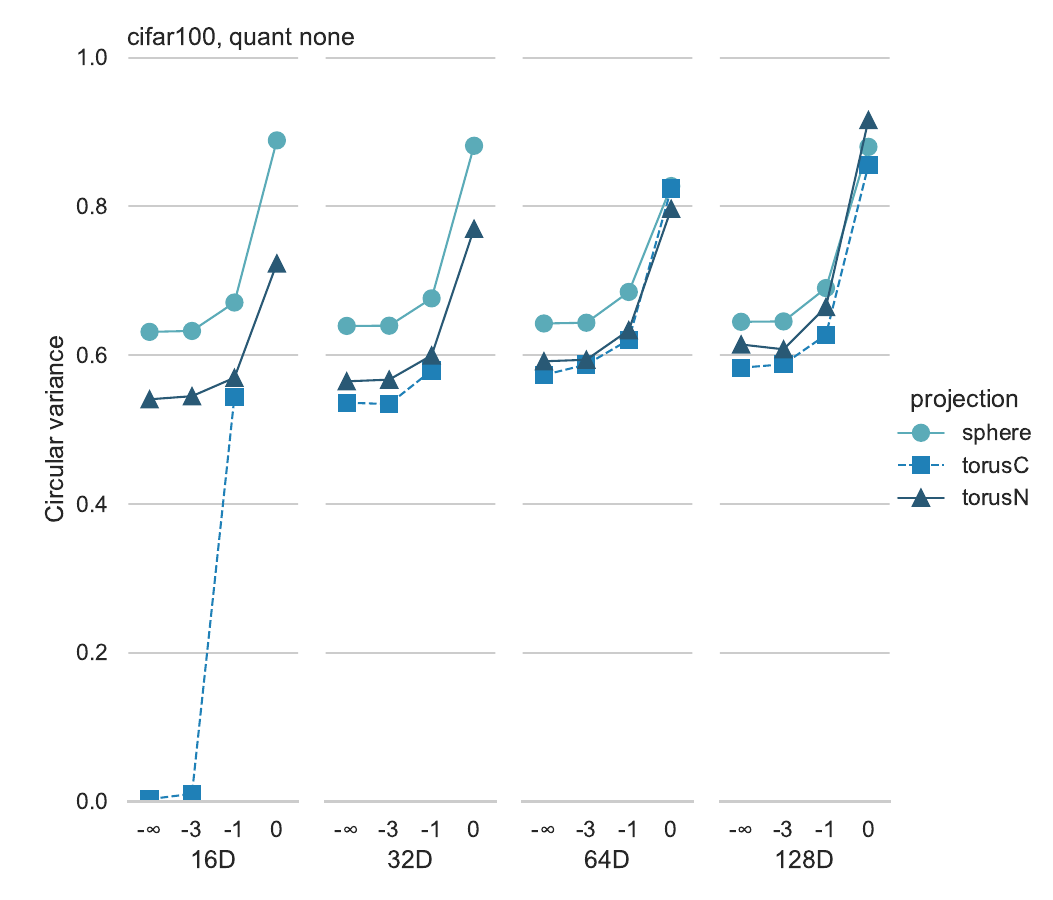}
  \includegraphics[width=0.31\columnwidth,page=9,clip,trim=0mm 0mm 0mm 0mm]{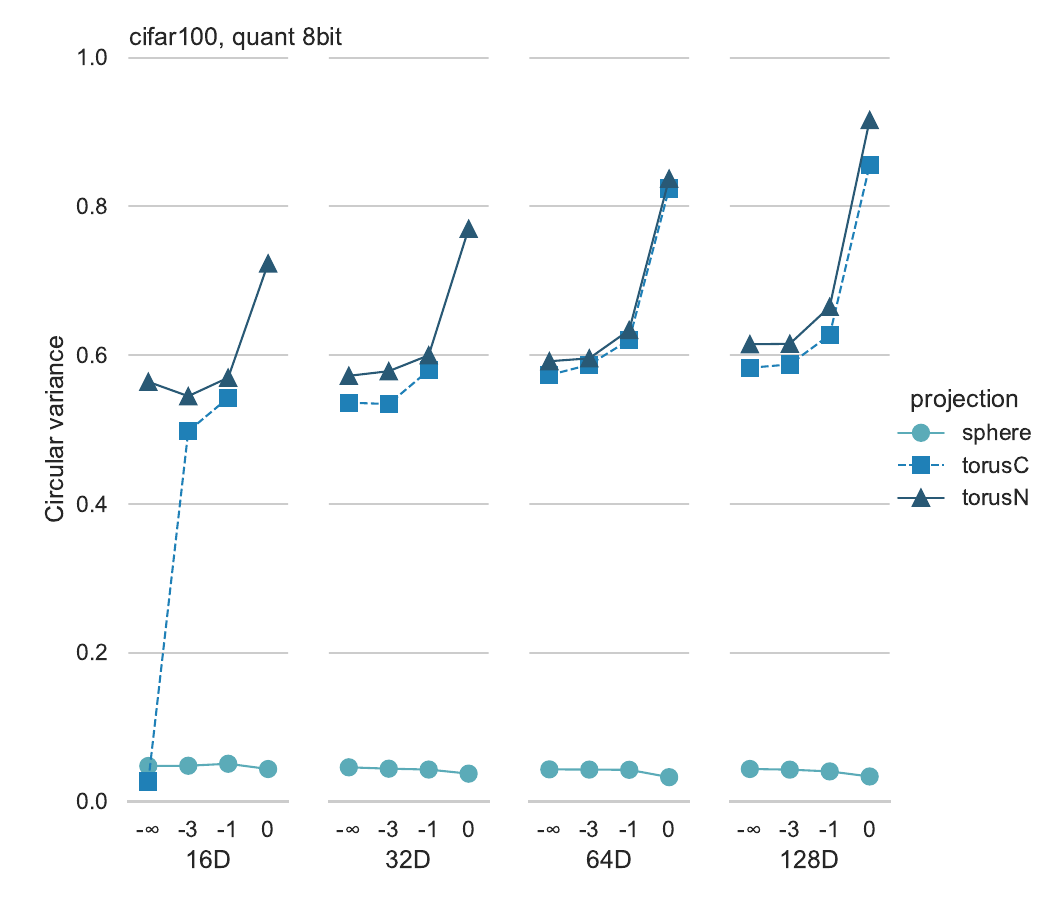}
  \includegraphics[width=0.31\columnwidth,page=10,clip,trim=0mm 0mm 0mm 0mm]{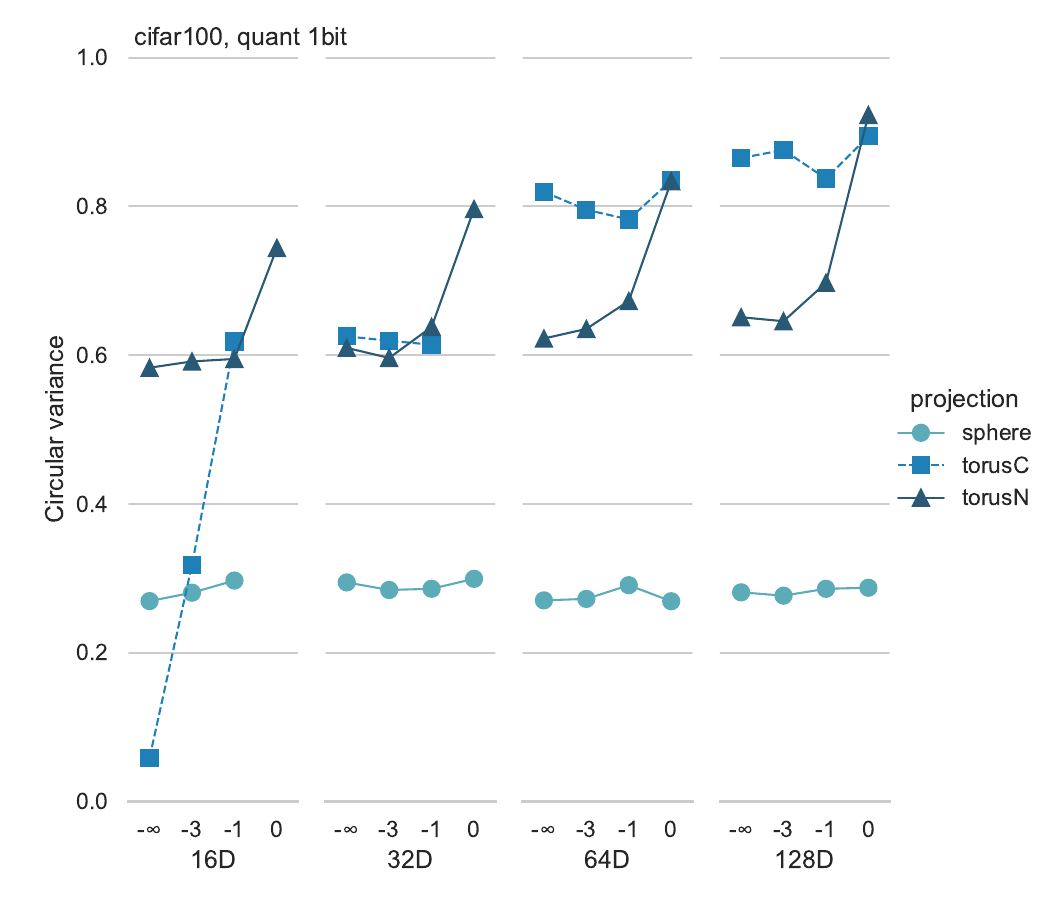}
  \includegraphics[width=0.31\columnwidth,page=12,clip,trim=0mm 0mm 0mm 0mm]{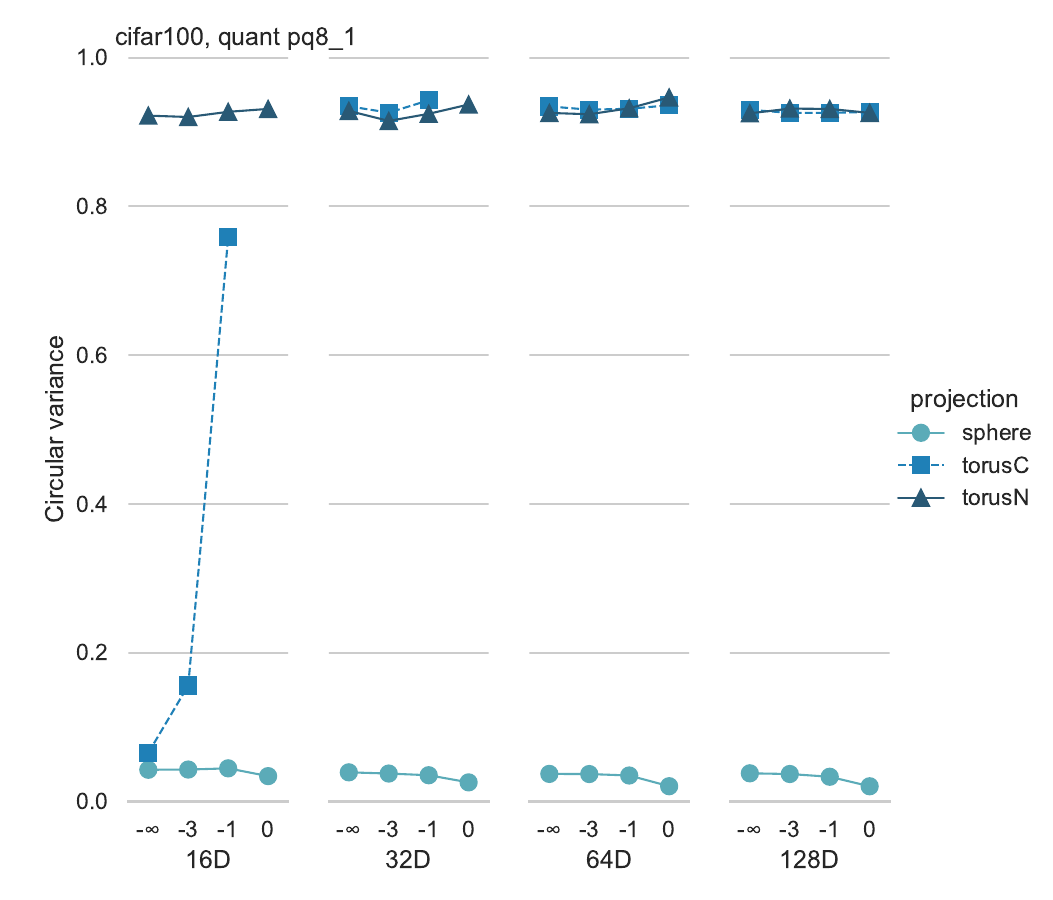}
  \includegraphics[width=0.31\columnwidth,page=13,clip,trim=0mm 0mm 0mm 0mm]{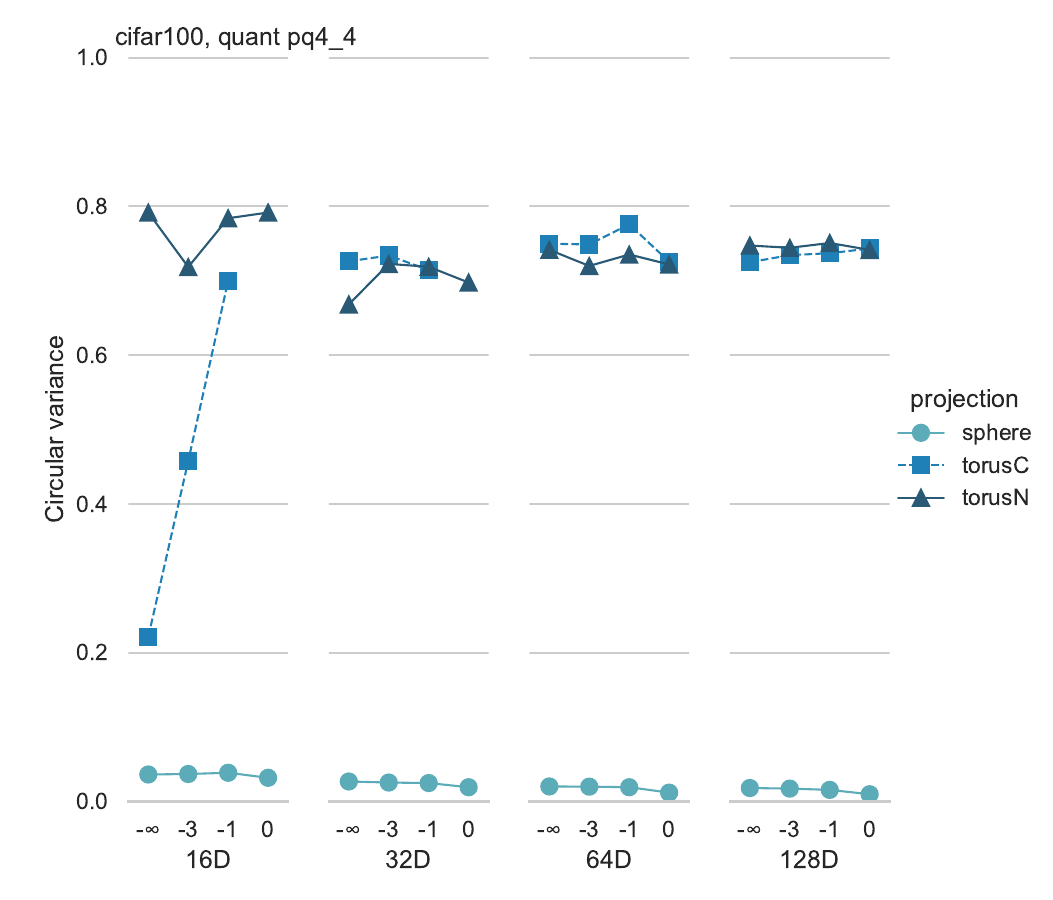}
  \includegraphics[width=0.31\columnwidth,page=14,clip,trim=0mm 0mm 0mm 0mm]{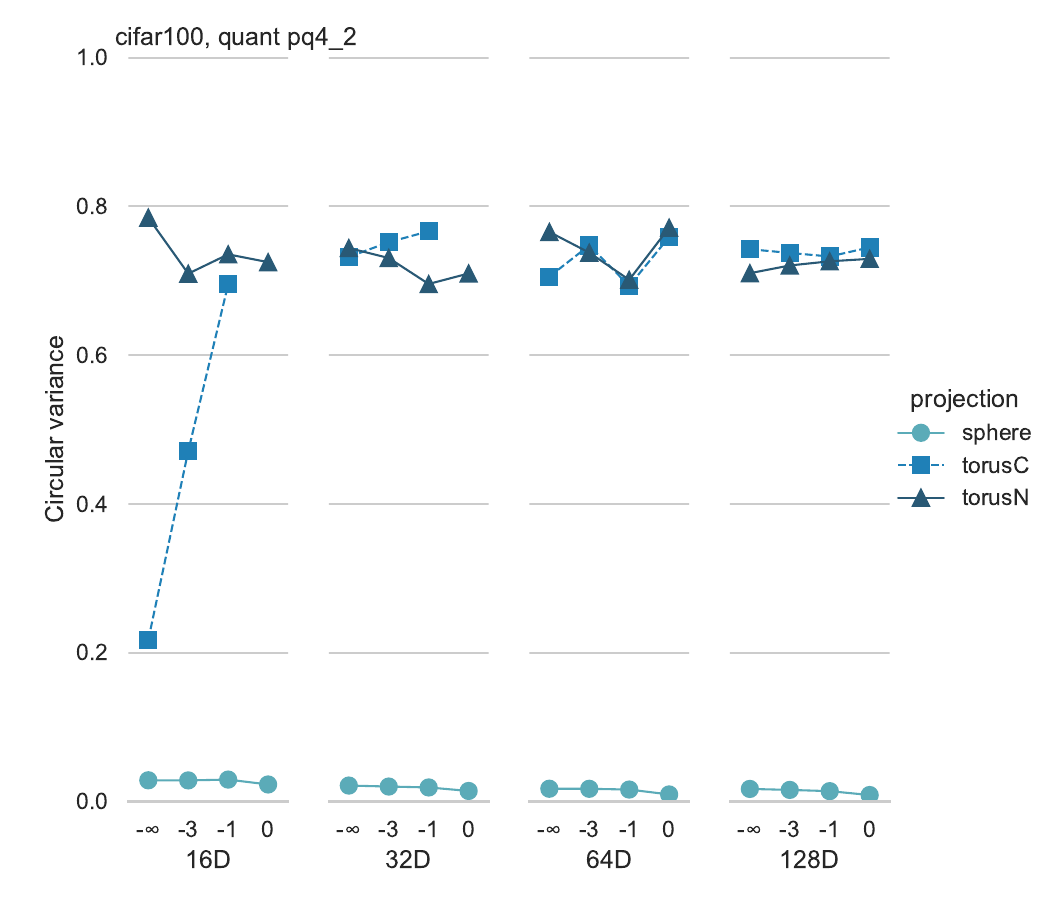}
  \caption{The circular variance of the representations evaluated in Figures \ref{fig:plot_evalmeasures_precision_at_1_raw} and \ref{fig:plot_evalmeasures_precision_at_1_quant} (CIFAR100 data set).}
  \label{fig:plot_evalmeasures_circvar}
\end{figure}

\end{appendices}

\end{document}